# Assessing Emoji Use in Modern Text Processing Tools 

**Abu Awal Md Shoeb**
Dept. of Computer Science
Rutgers University
New Brunswick, NJ, USA
`abu.shoeb@rutgers.edu`

**Gerard de Melo**
Hasso Plattner Institute /
University of Potsdam
Potsdam, Germany
`gdm@demelo.org`

## Abstract

Emojis have become ubiquitous in digital communication, due to their visual appeal as well as their ability to vividly convey human emotion, among other factors. The growing prominence of emojis in social media and other instant messaging also leads to an increased need for systems and tools to operate on text containing emojis. In this study, we assess this support by considering test sets of tweets with emojis, based on which we perform a series of experiments investigating the ability of prominent NLP and text processing tools to adequately process them. In particular, we consider tokenization, part-of-speech tagging, as well as sentiment analysis. Our findings show that many tools still have notable shortcomings when operating on text containing emojis.

## 1 Introduction

In our modern digital era, interpersonal communication often takes place via online channels such as instant messaging, email, social media, as well as websites and apps. Along with this growth of digital communication, there is an increasing need for tools that operate on the resulting digital data. For instance, online conversations can be invaluable sources of insights that reveal fine-grained consumer preferences with regard products, services, or businesses. Modern text processing and natural language processing (NLP) tools address a range of different tasks, encompassing both fundamental ones such as tokenization and part-of-speech tagging as well as semantic tasks such as sentiment and emotion analysis, text classification, and so on.

However, the shifts in modality and medium also shape the way we express ourselves, making it increasingly natural for us to embed emojis, images, hashtags into our conversations. In this paper, we focus specifically on emojis, which have recently become fairly ubiquitous in digital communication, with a 2017 study reporting 5 billion emojis being sent daily just on Facebook Messenger (Burge, 2017). Emojis are textual elements that are encoded as characters but rendered as small digital images or icons that can be used to express an idea or emotion.

**Goals.** Due to their increasing prominence, there is a growing need to properly handle emojis whenever one deals with textual data. In this study, we consider a set of popular text processing tools and empirically assess to what extent they support emojis.

Although emojis can be encoded as Unicode characters, there are special properties of emoji encoding that may need to be considered, such as skin tone modifiers, first introduced in 2015 for a small set of emojis. Some tools may handle regular emojis but fail to handle skin tones properly. Moreover, text harbouring emojis may adhere to subtly different conventions than more traditional forms of text, e.g., with regard to token and sentence boundaries. Finally, emojis may of course also alter the semantics of the text, which in turn may, for instance, affect its sentiment polarity.

**Overview.** For our analysis, we draw primarily on real tweets to study a range of different kinds of emojis. We run a series of experiments on this data evaluating each text processing tool to observe its behaviour at different stages in the text processing pipeline. Our study focuses on tokenization, part-of-speech tagging, and sentiment analysis. The results show that many tools have notable deficiencies in coping with modern emoji use in text.

## 2 Related Work

While emoji characters have a long history, they have quite substantially grown in popularity since their incorporation into Unicode 6.0 in 2010 followed by increasing support for them on mobile devices. Accordingly, numerous studies have sought to explain how the broad availability of emojis has affected human communication, considering grammatical, semantic, as well as pragmatic aspects (Kaye et al., 2017; McCulloch, 2019). Only few studies have specifically considered some of the more advanced technical possibilities that the Unicode standard affords, such as zero width joiners to express more complex concepts. For instance, with regard to emoji skin tone modifiers, Robertson et al. (2020) study in depth how the use of such modifiers varies on social media, including cases of users modulating their skintone, i.e., using a different tone than the one they usual pick.

Given the widespread use of emojis in everyday communication, it is important to consider their support in the most commonly used NLP toolkits, such as Stanford's Stanza (Qi et al., 2020) and NLTK (Bird et al., 2009), which power a wide range of applications. There are numerous reports that compare the pros and cons of popular NLP libraries (Wolff, 2020; Kozaczko, 2018; Choudhury, 2019; Bilyk, 2020). These primarily consider the features and popularity of the tools, as well as their performance. However, there have not been studies comparing them with regard to their ability to cope with modern emoji-laden text. Since emojis are becoming increasingly ubiquitous, it is crucial for developers and institutions deploying such software to know whether it can cope with the kinds of text that nowadays may quite likely arrive as input data. In many real-world settings, applications and services are expected to operate on text containing emojis, and so it is important to investigate these capabilities.

Many academic studies present new models for particular NLP tasks relating to emojis. For instance, Felbo et al. (2017) developed an emoji prediction model for tweets. Weerasooriya et al. (2016) discussed how to extract essential keywords from a tweet using NLP tools. Cohn et al. (2019) attempted to understand the use of emojis from a grammatical perspective, seeking to determine the parts-of-speech of emoji occurrences in a sentence or tweet. Owoputi et al. (2013) proposed an improved part-of-speech tagging model for online conversational text based on word clusters. Proisl (2018) proposed a part-of-speech tagger for German social media. However, these studies mostly target just one specific task and are typically not well-integrated with common open source toolkits.

## 3 Experimental Data

As we want to assess the support of emojis provided by different text processing tools, we first consider some of the different cases of emoji use that one may encounter, in order to compile relevant data.

### 3.1 Emoji Use in Text

Emojis can appear in a sentence or tweet in different ways. They may show up at the beginning of a tweet or at the end of a tweet. Similarly, they may appear as part of a series of emojis separated by spaces, or could be clustered within a tweet without any interleaved spacing. Based on observations on a collection of tweets crawled from Twitter (Shoeb et al., 2019), we defined a series of cases distinguishing different aspects of emoji use, including the number of emojis, position of emojis, the use of skin tone modifiers, and so on.

**Case 1: Single Emoji.** This is the simple case of emoji use with only one single emoji occurrence in the entire tweet. However, in this case, an emoji can be space-separated from the text or it can be tied up with text without having any leading or trailing spaces. This is the rudimentary case among all cases designed to quickly assess if there is bare minimum emoji support offered by the respective text processing tool.

| Case 1.1: Single Emoji with Space |
|---|
| Emojis 😀 are a new way of expressing emotions! #emoji |

| Case 1.2: Single Emoji without Space |
|---|
| Emojis😀are a new way of expressing emotions! #emoji |

**Case 2: Multiple Emojis.** In real-world social media, we often observe multiple emojis within a single posting. In this case, emojis may be found in multiple places as single emojis or as a group. Use of the same emojis repeatedly within a tweet is a common phenomenon, especially when people wish to emphasize or express a high intensity of an emotion. Such use is akin to the repetition of individual characters in OOV words such as *funnnn*, *heloooo*, which may also be encountered. The two may also occur together: For example, when people write "Yahooooo!" to express excitement, they may also be likely to add multiple emojis " 💃💃💃 " instead of a single one.

> **Case 2.1: Multi Emoji Multi Positions**
> 
> Emojis 😀 are a new way for expressing emotions 😊! #emoji

> **Case 2.2: Multi Emoji with Space**
> 
> Another example is having multiple emojis 😀 😀 😀 😀 together in a tweet.

> **Case 2.3: Multi Emoji Cluster**
> 
> This gets a little complicated when having multiple emojis 😀😀😀😀 in a tweet without having any spaces in between emojis.

**Case 3: Emojis with Skin Tone Modifiers.** Skin tone modifiers, introduced in Unicode 8.0, allow users to modify the skin color tone in an emoji to their liking, enabling a more accurate representation of diverse appearances. The Unicode standard adopts the Fitzpatrick Scale (Fitzpatrick, 1975), according to which the skin tone for selected emojis can be modulated with five different color settings:

- 🏻 Light Skin Tone (e.g. 👍🏻)
- 🏼 Medium-Light Skin Tone (e.g. 👍🏼)
- 🏽 Medium Skin Tone (e.g. 👍🏽)
- 🏾 Medium-Dark Skin Tone (e.g. 👍🏾)
- 🏿 Dark Skin Tone (e.g. 👍🏿)

Internally, an Emoji Modifier Sequence is created automatically when a modifier character follows a supported base emoji character, resulting in a single emoji with skin tone.

> **Case 3.1: Skin Tone Emoji with Space**
> 
> I'm the Face with Tears of Joy emoji 😂. How do you like 👍🏻 me?

> **Case 3.2: Skin Tone Emoji All Colors**
> 
> We are all same 👶 👶🏻 👶🏽 👶🏾 👶🏿 but different in skin colors!

> **Case 3.3: Skin Tone Emoji Long Sequence**
> 
> Checking a long sequence of emojis 😂😂😂😂😂😂😂 and skin tones 👋👋🏻👋🏼👋🏽👋🏾👋🏿.

| Tweets | Count | % |
|---|---|---|
| Total | 22.3 M | 100 |
| Unique | 21.4 M | 95.84 |
| Only single emoji | 5.67 M | 25.38 |
| Multiple emojis | 16.48 M | 73.77 |
| Emoji skin tone modifiers | 1.31 M | 5.85 |
| Light Skin Tone emojis | 382 K | 1.71 |
| Medium Light Skin Tone emojis | 386 K | 1.73 |
| Medium Skin Tone emojis | 337 K | 1.51 |
| Medium Dark Skin Tone emojis | 274 K | 1.23 |
| Dark Skin Tone emojis | 53 K | 0.24 |
| Zero Width Joiner (ZWJ) emojis | 97 K | 0.43 |

Table 1: Corpus statistics – the distribution of emojis over the ~22 million tweets with regard to the considered cases

Not all software supports this more recent addition to the Unicode standard. For example, some software may fail to render such emojis. In our assessment, we do not consider rendering aspects, but wish to ascertain that the modifier sequence remains intact, rather than treating a skin tone modified emoji as two separate characters.

**Case 4 & 5: Emojis from Basic and Supplemental Planes.** In the Unicode standard, a plane is a continuous group of different code points (Wikipedia contributors, 2020). Some characters now classified emojis are encoded in Plane 0, the Basic Multilingual Plane, where 16 bits suffice to encode individual characters. However, the majority of the emojis reside in Plane 1, the Supplementary Multilingual Plane, which in the past had mainly been reserved for rare historic scripts. When including the latter, individual characters can no longer be encoded directly within just 16 bits. Hence, we consider whether a tool handles both non-BMP and BMP emojis.

**Case 6: Emojis with Zero Width Joiner (ZWJ).** Zero Width Joiners, pronounced "zwidge", are not emojis per se, but rather join two or more other characters together in sequence to create a new one. Popular emoji ZWJ sequences include group ones such as 👨‍👩‍👧‍👦 the Family: Man, Woman, Girl, Boy emoji, which combines 👨 Man, the U+200D ZWJ code, 👩 Woman, U+200D again, 👧 Girl, U+200D, and finally 👦 Boy. These are rendered as a single emoji on supported platforms.

### 3.2 Tweet Selection

Given the different cases of emoji use discussed above, we search for relevant examples in the Twitter corpus. Table 1 provides corresponding statis-tics, showing that even rare phenomena do occur in substantial numbers of tweets.

Next, we chose representative samples for each case. We restricted our search to English language tweets. and made sure that not all tweets simply consisted of URLs or mentions. The latter is fairly common on Twitter, and since it would not be very uncommon for a text processing tool to encounter such a tweet, we did also incorporate a few such tweets along with tweets containing proper text. We also attempted to emphasize the motivation for the different cases given in Section 3.1. For example, we ensured that there are no modified emoji or skin tone modifiers among the examples for Cases 1 and 2. Since these are the first and most elementary tests, it appears desirable to first observe if a tool is able to deal with the most basic forms of emoji.

> An Example from Twitter
>
> When armed with this everything gets a clean 😂😂😂😂 including the neighbours car 🚗👍😂 ooops #foambath #jetwashing-fun #happydays #everythingclean

We attempted to maintain the same tweet selection criteria for all tasks in our experiments. The following sections describe the three considered major tasks, i.e., Tokenization (Section 4), Part-of-Speech Tagging (Section 5), and Sentiment Analysis (Section 6) separately.

## 4 Tokenization

Tokenization is the act of breaking up a sequence of strings into a sequence of basic pieces such as words, keywords, phrases, symbols, and other elements, referred to as tokens. In the process of tokenization, some characters such as punctuation marks may be discarded. It is important for a tokenizer to generate meaningful results, as the output of this step becomes the input for subsequent processing steps such as parsing and text mining in the pipeline. In our study, we expect a tokenizer to segment a text into tokens such as words, emojis, and other special characters.

### 4.1 Task Setup

While tokenizing a sentence, or a tweet with emojis, in particular, we considered the cases presented earlier in Section 3, considering in particular how well emojis can be separated from words. An emoji can accompany a word with both leading and trailing spaces, or it can be concatenated with words without any additional whitespace. We expect a tokenizer to distinguish an emoji from a word even in the absence of a space delimiter. The same principle should be followed for emoji clusters, i.e., if multiple emojis occur in a sequence such as "🎂🥂🎈", they are expected to be treated as individual tokens.

Another aspect of successful tokenization is adequately handling emoji skin tone modifiers. As we have five different skin tones, we ensure that our test data for Case 3 contains the same number of tweets from all skin tones. An ideal tokenizer should not split skin tone emoji into two individual characters. For example, the *Waving Hand Light Skin Tone* 👋 emoji should not be split into a regular *Waving Hand* emoji 👋 and a tone modifier 🏻.

We also test the abilities of tools in terms of handling ZWJ emoji sequences. We randomly pick a small set of tweets containing ZWJ sequences for this purpose. For example, an ideal tokenizer should not split up a Family Emoji as four individual emojis such as Man, Woman, Girl, Boy, as the emoji is meant to be rendered as a single one.

Note that some tokenizers discard punctuation during the tokenization process, while others retain them as tokens. For example, *Gensim* removes all punctuation, including all emojis. Similarly, the *NLTK Tweet Tokenizer* does not split up a hashtag as # followed by a word, but rather keeps the hashtag intact, as hashtags usually convey meaningful information in tweets. Therefore, to generalize the tokenization process for all tools, we apply some post-processing techniques before comparing the list of tokens with the expected list of tokens. One such technique is to discard all punctuation from the list of tokens. However, for #hashtag occurrences, we treat both "hashtag" and "#hashtag" as correct tokens.

**Tools.** In total, we consider 8 libraries for our experiments. These are the regular English tokenizer of the Natural Language Toolkit (NLTK) by Bird et al. (2009), the NLTK Tweet Tokenizer (i.e., its Twitter-aware tokenizer), the Stanford NLP Group's Stanza (formerly known as StanfordNLP) (Qi et al., 2020), SpaCy and SpaCyMoji, PyNLPl (the Python library for Natural Language Processing, pronounced as *pineapple*), Gensim (Řehůřek and Sojka, 2010), TextBlob, and AllenNLP (Gard-ner et al., 2018).

| Tools | Task - Tokenization | | | | | |
|---|---|---|---|---|---|---|
| | Case 1 Single Emoji | Case 2 Multi Emoji | Case 3 Skin Tone Emoji | Case 4 BMP Plane 0 | Case 5 non-BMP Other Planes | Case 6 Zero Width Joiner |
| Gensim | 0 | 0 | 0 | 0 | 0 | 0 |
| NLTK | 80 | 0 | 68 | 40 | 70 | 70 |
| NLTK-TT | 70 | 70 | 0 | 80 | 60 | 0 |
| PyNLPl | 50 | 0 | 38 | 30 | 20 | 70 |
| SpaCy | 100 | 100 | 0 | 100 | 100 | 0 |
| SpaCyMoji | 100 | 100 | 92 | 100 | 100 | 0 |
| Stanza | 90 | 10 | 68 | 50 | 90 | 40 |
| TextBlob | 80 | 0 | 68 | 40 | 70 | 70 |

Table 2: The percentage of success of tools covering all different cases of emojis in tokenization

### 4.2 Results

Table 2 presents the results of tokenizing the given case-specific test data, based on an overall set of 100 real tweets. We partitioned this test data with regard to Cases 1 to 6 for a more fine-grained analysis.

For Case 1, intended to be the simplest one, where each input cannot contain more than one emoji, the test set consists of 10 tweets. Here, we observe that most tools except for *Gensim* partially pass this test case. Since *Gensim* discards emoji characters, it also fails all other test cases. In contrast, both *SpaCy* and *SpaCyMoji* achieve 100 percent accuracy for tokenizing tweets with a single emoji. Other tools may fail to segment off emojis that have been attached to words without whitespace.

Case 2 is designed for multiple emojis, including clusters of emojis, for which we also have 10 test cases. Some tools, including *NLTK* and *PyNLPl*, failed for this part despite having done well on Case 1. Apart from separating off emojis from words, tools here differ mostly based on whether they split up groups of emojis.

For Case 3, there are 50 test cases with skin tones. Note that these can have single or multiple emojis, but it is ensured that they bear at least one skin tone emoji. In some cases, the problems are the same as for regular emojis, e.g., splitting off emojis from words. However, some tools generally split off skin tone modifiers from the emojis they are intended to modify. *Stanza* only breaks a color tone emoji into the base emojis and tone modifiers when it is concatenated with text. Otherwise it can handle a skin tone emoji without splitting it. *SpaCyMoji* obtains a near-perfect result but still does not manage to preserve all skintone emojis.

Cases 4 and 5 are designed for handling Basic Multilingual Plane (BMP) and non-BMP emojis, respectively. For each of these cases, a distinct set of 10 tweets was used to assess the performance. Interestingly, non-BMP emojis appear to be better-supported, presumably because they include the most popular emojis.

Finally, Case 6 considers emojis with zero width joiners (ZJW). This test set contains 10 tweets in total, where each tweet contains no more than two emojis with at least one ZWJ emoji. The tools that fail in this case, such as *NLTK-TT*, instead of preserving a ZJW emoji such as 👨‍👩‍👧, produce multiple separate tokens, including the Unicode zero-width joiners as individual tokens, e.g., 👨, U+200D, 👩, U+200D, 👧, U+200D, and 👦. In fact, none of the tools could achieve 100% accuracy across all ZWJ emojis. This is because they may fail when a regular emoji and a ZWJ one appear together. For example, one of tweets contains 🔒👨‍👩‍👧 emojis and *NLTK* treats them as one token, although it successfully handles other ZWJ emojis when they are space-separated. In contrast, *NLTK-TT* appears to be the best option for dealing with emoji clusters, but when it comes to ZWJ emojis, it separates all emojis and joiners.

## 5 Part-of-Speech Tagging

Part-of-Speech (POS) tagging is the process of assigning each token a label that reflects its word class. This may be with respect to traditional parts of speech, such as noun, verb, adjective, etc., or using a more fine-grained inventory of classes.

|  | *Task - Parts-of-Speech (POS) Tagging* | | | | | | |
|---|---|---|---|---|---|---|---|
| **Tools** | **Noun 26%** | **Adjective 22%** | **Verb ~17.3%** | **Adverb ~17.3%** | **Punctuation ~17.3%** | **Average 100%** | **Modified Tokenizer** |
| NLTK | 100.0 | 0 | 0 | 0 | 0 | 26.1 | 26.1 |
| NLTK-TT | 83.3 | 100 | 100 | 0 | 0 | 60.9 | 60.9 |
| SpaCy | 66.7 | 0 | 100 | 0 | 0 | 34.8 | 34.8 |
| SpaCyMoji | 66.7 | 0 | 100 | 0 | 0 | 34.8 | 34.8 |
| Stanza | 83.3 | 20 | 100 | 25 | 0 | 47.8 | ↑ 52.2 |
| TextBlob | 83.3 | 20 | 100 | 0 | 0 | 43.5 | ↑ 60.9 |

Table 3: The percentage of success of tools at labeling emojis with different parts-of-speech. The last column reports the average percentage of success when a modified tokenizer is used.

| Tools | Tweets | Target Emoji | Expected POS | Default Tokenizer | Modified Tokenizer |
|---|---|---|---|---|---|
| Stanza | She kept her 🐕 dog but had to sell her 🐱…. | 🐕 | Noun | 🐕 (ADJ) 🐱…. (.) | 🐕 (ADJ) 🐱 (NN) |
| Stanza | I MADE A PICTURE ‼️ ‼️ What do you think ❓ ✨😱 | ‼️ | Punctuation | ‼️ ‼️ (.) ❓ ✨ (NNP) | ‼️ (NN) ‼️ (.) |
| TextBlob | I MADE A PICTURE ‼️ ‼️ What do you think ❓ ✨😱 | ❓ | Punctuation | ❓ ✨😱 (NNS) | ❓ (NNP) ✨ (NNP) 😱 (NN) |
| TextBlob | Yes, she is😎and I like it | 😎 | Adjective | is😎and (Verb) | 😎 (Adj) |
| Stanza | **MODIFIED:** She kept her 🐕 but had to sell her 🐱…. | 🐕 | Noun | 🐕 (Noun) 🐱…. (.) | 🐕 (Noun) 🐱 (Noun) |

Table 4: Examples of tweets in which an emoji assumes the role of different parts-of-speech. The last column reports how the tagging accuracy can be improved by utilizing a unified tweet-aware tokenizer across all tools.

### 5.1 Task Setup

In order to understand how different POS taggers handle emojis in a sentence, we evaluate all tools for a subset of the given cases listed in Section 3.

For evaluation, we compiled a set of 23 real tweets from the same emoji corpus, in which emojis are used as different parts-of-speech, namely as nouns, adjectives, verbs, adverbs, or as punctuation. We mapped the original part-of-speech tags to these coarse-grained categories and then checked for correctness. Only the part-of-speech tags assigned to the emojis was considered, while the tagging of all other non-emoji tokens was disregarded in this experiment.

**Tools.** For this task, we evaluated all tools except *Gensim* and *PyNLPl*, as they do not directly offer any POS tagging functionality. Since tokenization is a prerequisite for POS tagging, a tool is likely to fail to correctly tag a word or emoji if the emoji is not properly tokenized in the preceding step. However, for a more extensive evaluation, we considered two setups. First, we conducted the POS tagging experiment based on the output of the integrated tokenizer of the respective tools. Thus, if a tool was unable to tokenize "Emojis😀are" as three separate tokens "Emojis", "😀", and "are", we still proceeded with the task treating it as one token for the respective tool's POS tagger. Subsequently, we conducted the POS tagging experiment while considering a unified tokenization as input for all tools. For example, in the case of "Emojis😀are", the tagger could expect to receive them as separate tokens "Emojis", "😀", and "are".

### 5.2 Results

Table 3 provides the results for our part-of-speech tagging experiments. The final two columns summarize the results with the original tokenizer and the modified tokenizer. None of the tools in our experiment could handle the case of emojis acting as adverbs or as punctuation. For instance, "My Credit Score Went 🔺 7 Points 🙏"is one such example where the *Upwards Button* 🔺 emoji assumes an adverbial role, which none of the taggers recognize, despite the emoji being space-delimited.

Similarly, occurrences of the question mark emoji

| Tools | Model | NS | Emojis +ve | -ve |
|---|---|---|---|---|
| NLTK | VADER | 100.0 | 0.0 | 0.0 |
| TextBlob | PatternAnalyzer | 100.0 | 0.0 | 0.0 |
| TextBlob | NaiveBayesAnalyzer | 100.0 | 0.0 | 0.0 |

Table 5: Accuracy (in percentage) of different tools at predicting sentiment scores of neutral sentences alone (NS) or neutral sentences along with positive (+ve) or negative (-ve) emojis

❓ or double exclamation mark emoji ‼️ used as punctuation are marked as nouns by all tools.

Another interesting trend we can notice from Table 3 is the 100% success rate for handling a verb emoji except for the *NLTK* tool. Although *NLTK* is the only tool that passes all test cases for noun emojis, it fails for all other cases. Overall, *Stanza* and *TextBlob* obtain the highest success rate reported in the second last column of the table.

When considering the harmonized ground truth tokenization, as reported in the final column of Table 3, the results for *TextBlob* are boosted significantly and for *Stanza* a more modest gain is observed. *TextBlob* and *Stanza* for instance may fail when emojis are not separated by whitespace from regular words (e.g., love❤️) or from another emoji (e.g., 🤔🤓). Adding whitespace between the word and the emoji improves the results for both of them.

The first example in Table 4 shows the interesting phenomenon of redundancy causing incorrect predictions. In this tweet, both the dog emoji 🐕 and the cat emoji 🐱 are expected to be tagged as nouns, but Stanza assumes the former to be an adjective due to the additional presence of the regular word "dog". To examine this further, we also considered several modifications of the original tweet. First, we considered the tweet without the additional word "dog" word after the dog emoji 🐕, in which case Stanza can easily identify it as a noun. This is reported in the last row of Table 4. We also tried replacing the dog emoji with the word "dog" to see if *Stanza* can cope with erroneous word reduplication, and it turned out that Stanza could correctly identify both occurrences as nouns. Finally, we considered replacing the word "dog" with another 🐕 emoji. In this case, the tool marked the first 🐕 as a noun and the second 🐕 as punctuation.

## 6 Sentiment Analysis

Although the word "emoji" is not etymologically related to the word "emotion", numerous studies show how emojis can help to express emotions and sentiment in textual communication (Novak et al., 2015). Keeping this in mind, we also assessed how well NLP tools fare at the task of predicting the sentiment polarity of a text harbouring emojis. Table 6 shows examples of texts with different emojis. While the text alone may be ambiguous with respect to its sentiment polarity, the emoji appears to eliminate much of the ambiguity. For our study, the sentiment of emojis was determined based on the data by Novak et al. (2015) and the emojis were appended to the end of a sentence. The goal of this endeavour is to examine if the sentiment polarity is predicted correctly when a high-intensity emoji is incorporated into a neutral sentence.

### 6.1 Task Setup

For this task, we considered a set of texts with neutral or ambiguous sentiment. Each example was then modified with both positive and negative emojis, giving us the opportunity to observe whether the predicted polarity of the original tweet changes in accordance with the polarity of the emojis. For example, *I'll explain it later* is a neutral sentence that is modified either with a positive emoji 😊 or with a negative one such as 👎. We use different sets of positive and negative emojis to modify the sentiment of the text, covering a broad spectrum of the sentiment polarity of emojis.

**Tools.** Although many tools could be trained on labeled set of tweets, we sought to assess pre-existing systems as they are often used out-of-the-box without additional training or fine-tuning. Hence, this study considers *NLTK* and *TextBlob*, as they can be used on the fly without any training. *TextBlob*'s Sentiment module contains two sentiment analyzers (PatternAnalyzer and NaiveBayesAnalyzer, the latter trained on movie reviews). For *NLTK*, we use VADER (Valence Aware Dictionary and sEntiment Reasoner), a lexicon and rule-based sentiment analysis tool that is specifically attuned to sentiment as expressed in social media.

### 6.2 Results

The results are given in Table 5. In the sentiment prediction task for a given tweet with emojis, neither of the considered tools appears to be able to consider the emojis as part of their sentiment polarity prediction. Although VADER – used in *NLTK* – has basic emoji support, it does not appear to work well enough, as it is based on emoji definition strings. Although we apply different combinations of sentiment emojis to all inputs, the overall sentiment does not change for either of the tools. Some studies (Jain et al., 2019) show that an emoji can moderate the sentiment of a given tweet if the sentiment of an emoji is considered during training. Clearly, systems trained on emoji-bearing data can learn to consider them during prediction if their tokenization is handled properly and they are not discarded during preprocessing. However, given the importance of emojis in conveying sentiment, it appears that most out-of-the-box tools ought to consider emojis as well.

| Sentences | Sentiment Predictions | | |
|---|---|---|---|
| | Only Text | Only Emoji | Text +Emoji |
| They decided to release it 👎 | Neutral | –ve | –ve |
| They decided to release it 😒 | Neutral | –ve | –ve |
| Let's go for it 💃 | Neutral | +ve | +ve |
| My driver license is expired by little over a month 😟 | –ve | –ve | –ve |
| They are going to start a direct flight soon 😢 | Neutral | –ve | –ve |
| They are going to start a direct flight soon 😍 | Neutral | +ve | +ve |
| I'll explain it later 😍 | Neutral | +ve | +ve |

Table 6: Example sentences with emojis that can moderate the overall sentiment of the sentence

## 7 Discussion

Overall, based on Table 8, we can see that there is not a single tool that perfectly handles all considered cases of emojis. Indeed, many text preprocessing modules routinely discard emojis along with punctuation characters as non-standard characters. Gensim by default follows this common approach, which may be suboptimal. *NLTK-TT* as well as Stanza help keep track of hashtags as they retain them with the "#" sign intact, whereas other tools split them up as two individual tokens or remove the "#". NLTK, Stanza, and *TextBlob* fail to tokenize emojis if emojis are tied up with other words, while spaCy, spaCyMoji, and *NLTK-TT* handle such cases. Note that accurate tokenization, e.g., splitting off emojis attached to words, can also be a prerequisite for many downstream tasks, such as enabling higher-quality text classification and information retrieval.

For POS tagging, somewhat surprisingly, almost all tools did well with verbs, while they all struggled with punctuation emojis as well as adverbs. The results for adjectives were as well quite mixed. Overall, *NLTK-TT* and *TextBlob* achieved the highest success rate for POS tagging, although both still struggle with adverbs and punctuation, which can also lead to adverse effects in downstream tasks such as syntactic parsing. Moreover, *TextBlob* requires the use of a modified tokenizer. Table 4 illustrates some real tweets and emojis associated with the respective POS. The example in the last row shows how *TextBlob* fails to label 😎 correctly because of its tokenization issues. However, the same task can be passed, reported in the last column of the same row, if an emoji-aware tokenizer is invoked.

Thus, in practice one may wish to consider a mix-and-match approach, using a tokenizer from one library and a tagger from another, or adding a post-processing step to modify the original tokenization before invoking additional models.

**Semantic Associations.** Finally, we also inspected semantic associations for particular kinds of emojis. We considered a 300-dimensional word2vec SGNS model trained on the EmoTag (Shoeb et al., 2019) dataset, and generated a set of nearest neighbours for selected target emojis.

Table 7 reports the nearest emoji neighbours for different skin tone variants of the *Clapping Hand* emoji. We observed that most of the top 5 neighbours for each emoji came from the same skin tone color except one for *Medium Light* and *Medium* tone emojis reported in Row 4 and 5, respectively. It appears that speakers who use skin tone modifiers frequently also use additional emojis that support such modification and that they naturally tend to use the respective modifier fairly consistently.

The last row of the same table shows the nearest neighbours for a ZWJ family emoji. Interestingly,

| Emoji | Nearest Neighbour Emojis |
|---|---|
| Clapping Hands (Regular) 👏 | 👌 👍 🙌 👊 👋 |
| Clapping Hands (Light) 👏🏻 | 🙌🏻 👌🏻 👍🏻 👊🏻 👋🏻 |
| Clapping Hands (Medium Light) 👏🏼 | 🙌🏼 👌🏼 👍🏼 [👏] 👋🏼 |
| Clapping Hands (Medium) 👏🏽 | 🙌🏽 👌🏽 👍🏽 [👏] 💪🏽 |
| Clapping Hands (Medium Dark) 👏🏾 | 🙌🏾 👌🏾 👍🏾 👊🏾 👋🏾 |
| Clapping Hands (Dark) 👏🏿 | 🙌🏿 👌🏿 👍🏿 👊🏿 👋🏿 |
| ZWJ Family (Man, Woman, Girl, Boy) 👨‍👩‍👧‍👦 | 👩‍👩‍👧 👩‍👧 👨‍👩‍👧 👨‍👧 👨‍👩‍👦 |

Table 7: Nearest neighbour (NN) emojis for the Clapping Hands and Family emojis. All nearest neighbours follow mostly the same color tone of the respective emojis except some indicated with [ ].

| Tools | SE | GE | STE | BMP | ZWJ |
|---|---|---|---|---|---|
| AllenNLP | ✓ | ✓ | ✗ | ✓ | ✗ |
| Gensim | ✗ | ✗ | ✗ | ✗ | ✗ |
| NLTK | ✓ | ✗ | ✓ | ✓ | ✓ |
| NLTK-TT | ✓ | ✓ | ✗ | ✓ | ✗ |
| PyNLPl | ✓ | ✗ | ✓ | ✓ | ✓ |
| SpaCy | ✓ | ✓ | ✗ | ✓ | ✗ |
| SpaCyMoji | ✓ | ✓ | ✓ | ✓ | ✗ |
| Stanza | ✓ | ✗ | ✓ | ✓ | ✗ |
| TextBlob | ✓ | ✗ | ✓ | ✓ | ✓ |

Table 8: An overview of popular text processing NLP tools and their emoji support. Single Emoji (SE), Group Emoji (GE), Skin Tone Emoji (STE), Basic Multilingual Plane (BMP) Plane 0 Emoji, Zero Width Joiner (ZWJ) Emoji

all of the nearest neighbours of this ZWJ emoji contain a ZWJ sequence as well.

## 8 Conclusion

Emojis have become an integral part of modern interpersonal communication and text encountered in chat messages, social media, or emails is often laden with emojis. Hence, it is important to endow NLP tools with emoji support not only to obtain a deeper understanding of this wealth of data but also to properly preserve and process them correctly in various text processing systems.

In this study, we assessed how well prominent NLP tools cope with text containing emoji characters. To this end, we evaluated a set of tools on three different tasks across a range of challenging test sets capturing particular phenomena and encodings.

Our study demonstrates that there are notable shortcomings in widely used NLP tools. Although many tools are partially capable of operating on emojis, none of them proved fully equipped to tackle the full set of aspects considered in our study. Hence, special care needs to be taken when developing applications that may encounter emojis.